\documentclass[11pt,a4paper]{article}
\usepackage[hyperref]{eacl2021}
\usepackage{times}
\usepackage{latexsym}

\usepackage{multirow}
\usepackage{subcaption}
\usepackage{amsmath}
\usepackage{amssymb}
\newcommand{\R}{\mathbb{R}}
\usepackage{graphicx}
\usepackage{csquotes}
\usepackage[utf8x]{inputenc}

\usepackage{rotating}
\usepackage{makecell, multirow}

\usepackage[skip=1ex]{caption}
\usepackage{booktabs}

\usepackage{microtype}

\aclfinalcopy
\def\aclpaperid{1164}

\title{Metric-Type Identification for Multi-Level Header Numerical Tables\\ in Scientific Papers}

\author{Lya Hulliyyatus Suadaa$^1$, Hidetaka Kamigaito$^1$, Manabu Okumura$^1$, Hiroya Takamura$^{1,2}$ \\
$^1$Tokyo Institute of Technology \\
$^2$National Institute of Advanced Industrial Science and Technology (AIST) \\
\texttt{lya@stis.ac.id}\\
\texttt{\{kamigaito,oku\}@lr.pi.titech.ac.jp}\\
\texttt{takamura@pi.titech.ac.jp}
}
\date{}

\begin{document}
\maketitle
\begin{abstract}

Numerical tables are widely used to present experimental results in scientific papers. For table understanding, a metric-type is essential to discriminate numbers in the tables. We introduce a new information extraction task, metric-type identification from multi-level header numerical tables, and provide a dataset extracted from scientific papers consisting of header tables, captions, and metric-types. We then propose two joint-learning neural classification and generation schemes featuring pointer-generator-based and BERT-based models. Our results show that the joint models can handle both in-header and out-of-header metric-type identification problems.

\end{abstract}

\section{Introduction}

\begin{figure}
    \begin{subfigure}[t]{0.3\textwidth}
    \includegraphics[page=1,width=0.95\textwidth]{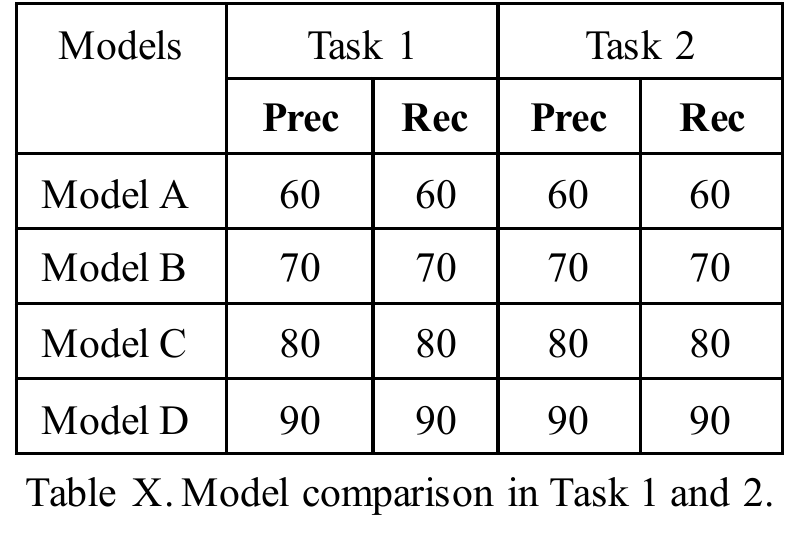}
    \caption{Metric-type in header} 
    \label{example-table-a}
    \end{subfigure}%
    \begin{subfigure}[t]{0.2\textwidth}
    \includegraphics[page=1,width=1\textwidth]{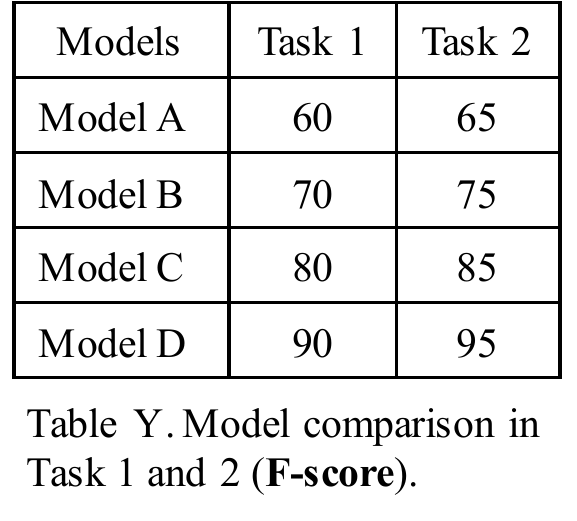}
    \caption{No metric-type in header} 
    \label{example-table-b}
    \end{subfigure}%
    \caption{Example tables in scientific papers. Bold indicates their metric-type.}
    \label{example-table}
\end{figure}

Tables are powerful tools for presenting data efficiently in row and column views. In scientific papers, numerical tables are commonly used to show experimental results for facilitating data analysis. Examples of numerical tables in scientific papers are shown in Figure \ref{example-table}.

Tables have the ability to cover multiple categories written in table headers by incorporating several header sets in a hierarchical view, called multi-level header tables. Scientific papers have strict guidelines about tables; for example, one states that a similar type of text is written in the same level of header. Figure \ref{example-table-a} shows a multi-level header example in the column part, with task type (\textit{Task 1} and \textit{Task 2}) in the first header-level and metric-type (\textit{Prec} and \textit{Rec}) in the second. The table also has a row header specifying the model type (\textit{Model A}, \textit{Model B}, \textit{Model C}, and \textit{Model D}). In the real-world, this header-type information is limited due to the unknown table scheme. However, we assume tables in scientific papers follow the rule of categorizing a similar type of header name in the same header-level.

To understand the numbers in the tables, metric-types are important for discriminating the numbers. A comparison between numbers is applied for numbers in the same metric-type with different categories. For the table in Figure \ref{example-table-a}, we cannot compare the number $60$ for \textit{Model A} in the first column with $60$ in the second one because they have a different metric-type: \textit{Prec} and \textit{Rec}. Computing numbers with different metric-types will result in inaccurate analysis.  

Different tables may have different ways of writing their header name, such as using abbreviations like \textit{p}, \textit{pre}, or \textit{prec} to refer to \textit{precision}. 
Due to the lexical diversity of header names, metric-type identification becomes more challenging. Using a rule-based metric-type tagging or a limited set of metric-types in a dictionary is not enough to cover the diversity. Since tables in scientific papers typically have logical captions and logical categorization of the header-level, we introduce a metric-type identification task that locates the metric-type in the headers by using the caption and header name as inputs. For the example shown in Figure \ref{example-table-a}, the metric-type is located in the second level of the column header.

We also cover tables that do not mention metric-types in their header (out-of-headers), as shown in Figure \ref{example-table-b}. In these cases, the metric-types are identified in the caption. To cover metric-types located both in the headers and not in the headers, we propose a joint framework of metric-type location prediction and metric-type token generation for the metric-type identification task in multi-level header tables.

Our contributions are as follows:
\begin{itemize}
\item We introduce a metric-type identification task for multi-level header tables and propose joint location prediction and generation models to solve the task.
\item We provide a dataset consisting of multi-level header numerical tables, captions, and metric-types, extracted from scientific papers. Our datasets will be publicly available\footnote{Dataset is available on \href{https://github.com/titech-nlp/metrictable}{https://github.com/titech-nlp/metrictable}}.
\item We introduce a multi-level header table encoder mechanism to obtain table header representations and propose a pointer-generator-based model to cover out-of-headers in the metric-type identification task.
\item We fine-tune a general pre-trained encoder (BERT) and a domain-specific encoder (SciBERT) in our task and present the experimental results. We show that the models incorporating the pre-trained encoders lead to significant performance gains, especially when using a domain-specific one.
\end{itemize}

\section{Related Work}

Table information extraction is beneficial to cover unknown table schemes and understand the table contents. 
\citet{milosevic2019} proposed a framework for table information extraction in biomedical domains by defining rules for all possible variables. Specifically, for numerical variables, they retrieved metric-types by searching a set of possible tokens in the dictionary. Focusing on numerical tables, \citet{Spread} extracted metric-types in earning reports by using similarity scores between the corresponding non-numeric text for the leftmost cells and stored metric-types.

The work closest to ours is the one by \citet{hou-etal-2019-identification}, who used tables from the experimental result section,  combined with the title and abstract as document representations to extract triples of tasks, dataset, and metric for leaderboard construction. In our study, we represent the tables in more generic ways, preventing the original table structure in the multi-level headers form. We intend to retain the ability of a table to cover complex categorization in the headers and efficiently present all values. A previous study that also explored multi-dimensional tables was done by \citet{milosevic2006} to automatically detect table structures from XML tables.

Our pointer-generator-based model in the metric-type generation scheme is inspired by the promising results of the pointer-generator network \cite{pointer-generator} in the summarization task. The network deals with the out-of-vocabulary issue by joint copying from source texts and generating from vocabularies.

Recent studies have shown that pre-trained encoders can be successfully fine-tuned for downstream NLP tasks, thus avoiding the need to train a new model from scratch. A pre-trained encoder BERT \cite{devlin-etal-2019-bert} was trained on the BooksCorpus (800M words) and Wikipedia (2,500M words). For better-contextualized representation in the scientific domain, \citet{beltagy-etal-2019-scibert} introduced a domain-specific BERT model, SciBERT, which was trained on 1.14M papers from Semantic Scholar. \citet{friedrich-etal-2020-sofc} implemented both BERT and SciBERT on their models to solve the information extraction task and achieved significant performance gains.

\section{Metric-Type Identification for Numerical Tables}

\subsection{Datasets}
\label{mt_datasets}
We automatically extracted tables from the PDF files of scientific papers in the computational linguistics domain using PDFMiner and Tabula as extraction tools and filtered only numerical tables related to experimental results using the keywords \textit{evaluation}, \textit{result}, \textit{comparison}, and \textit{performance}. We used papers from the ACL and EMNLP conferences (2016 to 2019) on the ACL Anthology website as data sources.

\begin{figure}
\includegraphics[width=0.45\textwidth]{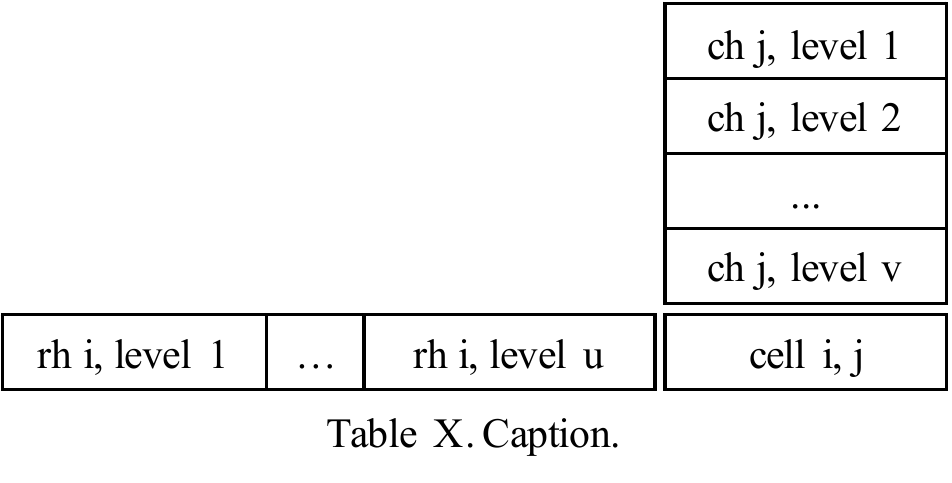}
    \caption{Table structure.}
    \label{table-structure}
\end{figure}

In tables in actual scientific papers, knowledge about the table semantics is rarely provided. On the basis of how information is \enquote{read} from a table, \citet{hurst_thesis} separated functional table areas into access cells and data cells. Access cells consist of column headers and/or row headers. We define data structure on the basis of their functional areas: table caption (capt), row headers (rh), column headers (ch), and cells. Headers in the row and column parts have several levels, and we assume that header names in the same level have the same type. Figure \ref{table-structure} shows our table structure.

\begin{figure}
\includegraphics[page=1,width=0.5\textwidth]{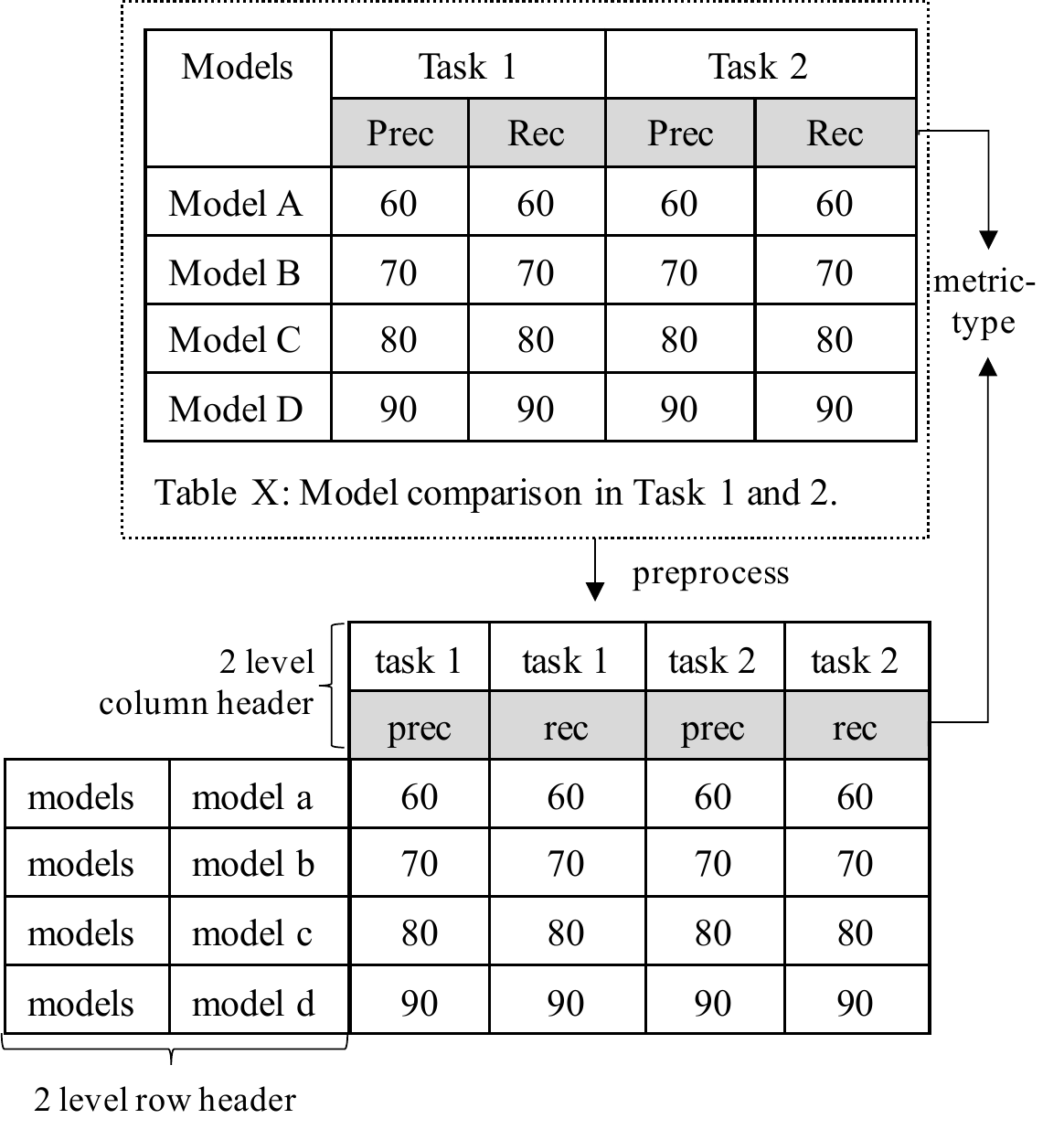}
    \caption{Illustration of table preprocessing.}
    \label{sample-preprocessing}
\end{figure}

We hired several qualified workers in the computer science field to manually check the extracted table structure to ensure the separation of row headers, column headers, and cells was correct, as shown in Figure \ref{sample-preprocessing}. Then, they annotated the metric-type of the tables by prioritizing the locating of the metric-type in a specific header-level. The annotators were able to identify the metric-types of approximately 70\% of the tables in their headers, and they determined the metric-type of the rest of the tables on the basis of information from the table captions. When no metric-type was mentioned in the headers, we assumed the metric-type was the same for all table values. The structures from the example in Figure \ref{sample-preprocessing} are capt: \enquote{{\it  model comparison in task 1 and 2}}; rh level 1: [{\it models, models, models, models}]; rh level 2: [{\it model a, model b, model c, model d}]; ch level 1: [{\it task 1, task 1, task 2, task 2}]; ch level 2: [{\it prec, rec, prec, rec}]; and metric-type: [{\it prec, rec, prec, rec}] (identified in ch level 2).

We split our dataset into training, validation, and test sets. The statistics of our dataset are provided in Table ~\ref{stat-table}.

\begin{table}[h]
\centering
\begin{tabular}{lrrr}
\hline \textbf{} & \textbf{Train} & \textbf{Val} & \textbf{Test} \\ \hline
No. of tables & 1,084 & 136 & 135 \\
Average row/column & 6 & 6 & 5 \\
Max level: & & \\
- row header & 9 & 6 & 4 \\
- column header & 6 & 5 & 6 \\
Vocab size: & & \\
- headers & 8,270 &1,435 & 1,230 \\
- all metric-types & 807 & 175 & 185 \\
- unique metric-types & 90 & 22 & 28 \\
\hline
\end{tabular}
\caption{\label{stat-table} Dataset statistics in training, validation, and test sets. }
\end{table}

\subsection{Problem Definition}
Let $\text{Table} = \{capt, rh^{i}_{k}, ch^{j}_{l}, cell_{ij}\} \text{, where } 1 \leq i \leq n_{r}, 1 \leq j \leq n_{c}, 1 \leq k \leq u, 1 \leq l \leq v$ denote an $n_{r} \times n_{c}$ table with the $u$ level of $rh$ and $v$ level of $ch$. The task is to identify metric-type set ($\hat{m}$) in the specific level of row header ($rh_{k}$) and column header ($ch_{l}$). To handle tables that do not include metric-types in their headers, we generate $\hat{m}$ by using information from the table caption. The formulation of the metric-type identification is as follows:
\begin{equation}
  \hat{m} =
    \begin{cases}
      \text{\{}rh^{i}_{k}\text{\}}^{n_{r}}_{i=1}, k \in \{1,...,u\} & \text{if $\hat{m}$ in $rh$}\\
      \text{\{}ch^{j}_{l}\text{\}}^{n_{c}}_{j=1}, l \in \{1,...,v\} & \text{if $\hat{m}$ in $ch$}\\
      \begin{aligned}[b]\text{\{}{w_{m}}\text{\}}_{\times j}, w_{m} \in W_{m} \text{ or }\\  w_{m} \in capt 
      \end{aligned}& \text{otherwise,}
    \end{cases}       
\end{equation}
where $W_{m}$ is a set of metric-types in the vocabulary.

\section{Models}
We propose neural models to identify the metric-type for multi-level header tables by means of a joint model of metric-type location prediction and metric-type token generation.

\subsection{Pointer-Generator with Supervised Attention Model}
\begin{figure*}
\centering
\includegraphics[page=1,width=0.85\textwidth]{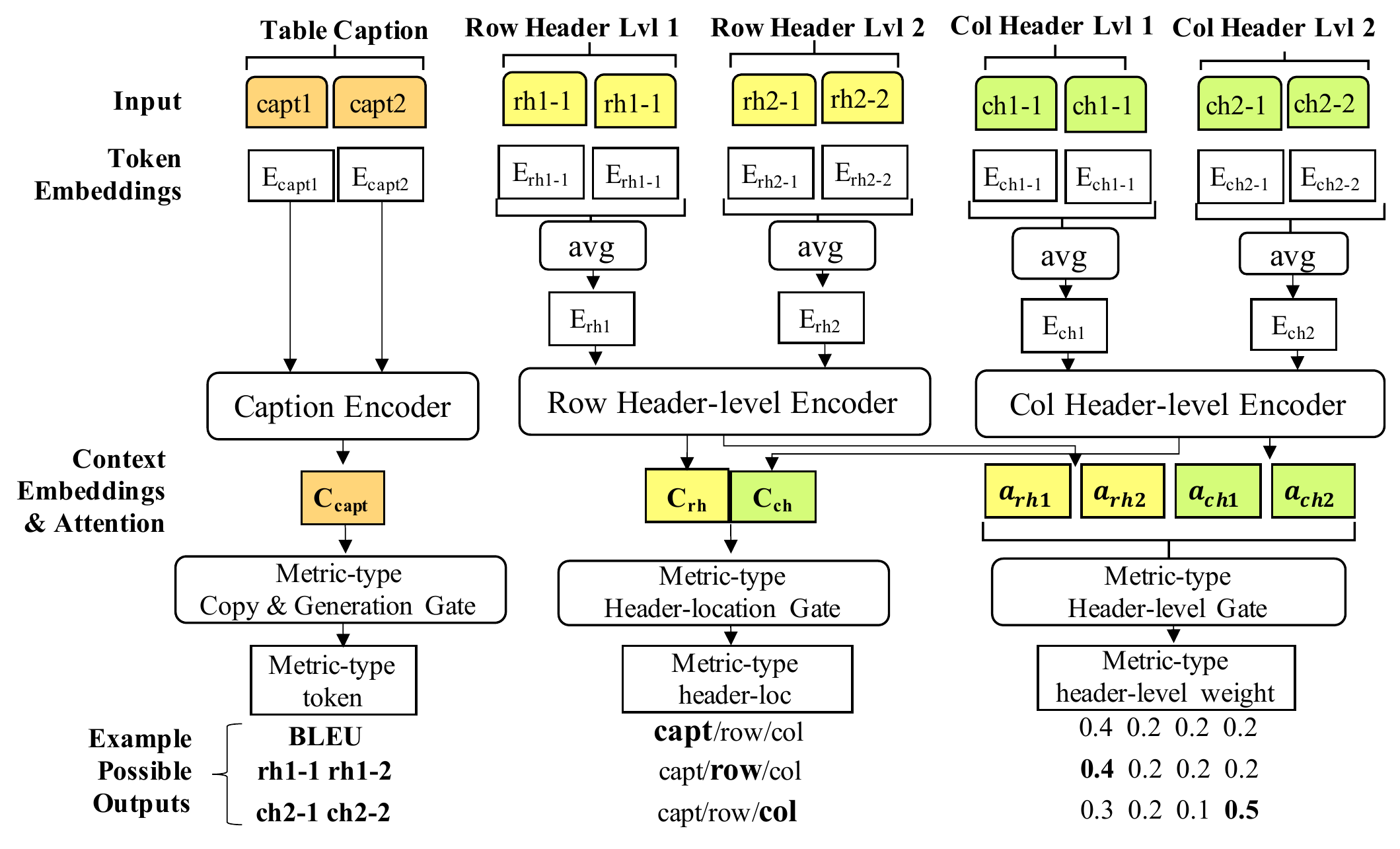}
    \caption{Architecture of proposed pointer-generator-based model to identify metric-types in tables.}
    \label{arch-pointergen}
\end{figure*}

We obtain the representations of captions and header-levels by using a BiLSTM encoder and then capture header-level weights using supervised attention between the header-level encoder and the metric-type header-location outputs. In the generation scheme, we adopt the pointer-generator network to take into account captions as source texts and the metric-type vocabulary in the metric-type generation gate. The architecture of our model is shown in Figure \ref{arch-pointergen}.

\paragraph{Header encoder} We use the vector representation of each header-level by averaging the vectors of all header name tokens in the same level. Given $E_{{rh}_{k}}$ and $E_{{ch}_{l}}$ as the averages of the initial vector representations of the row and column header-level vectors, respectively, we use the BiLSTM encoder with the dot attention mechanism proposed by \citet{luong-etal-2015-effective} to obtain the representations of the row and column header-levels and select the last hidden state of the last level combined with the weighted hidden states as header-level contexts, as follows:

\begin{equation}
C_{rh} = [C_{{rh}_{u}};\sum _{k=1}^{u}a_{{rh}_{k}}C_{{rh}_{k}}],
\end{equation}
\begin{equation}
C_{ch} = [C_{{ch}_{v}};\sum _{l=1}^{v}a_{{ch}_{l}}C_{{ch}_{l}}].
\end{equation}

\paragraph{Caption encoder} As with the headers, we use the BiLSTM encoder with attention  
$a_{capt_i}$ 
to compute the context vector of caption $C_{capt}$.

\paragraph{Metric-type header-location gates} We feed the concatenation of the row and column header contexts to the softmax layer to obtain the metric-type header-location probability:
\begin{equation}
p_{hloc} = \text{softmax}([C_{rh};C_{ch}]),
\end{equation}
which includes the probabilities of the metric-types located in row headers ($p_{rh}$), located in column headers ($p_{ch}$), or not located in the headers ($p_{capt}$), where $p_{rh}+p_{ch}+p_{capt}=1$.

\paragraph{Metric-type header-level gates} Since the attention scores $a_{{rh}_{k}}$ and $a_{{ch}_{l}}$ capture the relevant header-level information in row and column, these attention scores are used as header-level weights as follows:
\begin{equation}
\text{w}_{{hlevel}_{i}}= [a_{{rh}_{k}}p_{rh};a_{{ch}_{l}}p_{ch}],
\end{equation}
where $i\in\{1,...,u,(u+1),...,(u+v)\}$ as a header-level index.

\paragraph{Metric-type generation gates} In our pointer-generator network, we use the sigmoid layer to obtain a switch copy probability:
\begin{equation}
p_{copy} = \text{sigmoid}(C_{capt}),
\end{equation}
which lets us choose between copying word $w_{capt}$ from a table caption and generating word $w_{m}$ from the metric-type vocabulary, where $p_{copy} \in [0,1]$.
We use a softmax function to compute the probability distribution over the metric-type vocabulary:
\begin{equation}
P_{vocab}(w_{m}) = \text{softmax}(C_{capt}).
\end{equation}
Then, we obtain the following probability distribution over the extended vocabulary:
\begin{multline}
P(w_{m}) = p_{copy}\sum _{i:w_{i}=(w_{m})}^{n}a_{{capt}_{i}}  +\\ (1-p_{copy})P_{vocab}(w_{m}),
\end{multline}
where $i$ is the index of metric-type tokens in the vocabulary.

\paragraph{Learning objective} For training, we exploit the negative log-likelihood objective as the loss function. In addition, we adopt supervised attention \cite{liu-etal-2016-neural} for jointly supervising the row and column header-level attention to obtain the metric-type header-level. We combine all loss functions in the location classification and token generation model, and define $\alpha$ as the weight as follows:
\begin{multline}
\mathcal{L} = -((1-\alpha)(\sum _{c} z_{hloc} \log{p}_{hloc_{c}} +\\ \sum_{i=1}^{u+v}\log{\text{w}}_{{hlvl}_{i}}) + \alpha(\log{p}_{copy} +\\ \log P_{vocab}(w_{m}))),
\end{multline}
where $c \in  \{capt, rh, ch\}$ is the metric-type header-location classes and $z_{hloc}$ is the binary indicator (0 or 1) of each corresponding class.

\subsection{Fine-tuning BERT-based Model}

\paragraph{Input representation}
Input text in a fine-tuned BERT-based model is preprocessed by inserting two special tokens, [CLS] and [SEP]. In the original BERT architecture, [CLS] is appended to the beginning of input as the representation of the entire input sequence, and [SEP] is inserted after each input type as a sign of a segment boundary. For example, in question-answering tasks with two types of input text, pairs of question and answer, a
[CLS] token is appended before question tokens,
and [SEP] tokens are placed after question and after answer tokens, to separate the question and answer segments. Following \citet{liu-lapata-2019-text}, 
we customize these preprocessing schemes by inserting [CLS] before each segment and inserting [SEP] after each segment. We divide our inputs into several segments: caption, row header level $1$ to $u$, and column header level $1$ to $v$. 

The input text after preprocessing is denoted as a sequence of tokens $X = (x_{1}, x_{2}, \text{· · ·} , x_{n})$. 
There are three kinds of embedding assigned to each $x_{i}$: token embeddings representing the meaning of each token, segmentation embeddings indicating the segment boundaries of a sequence of tokens, and position embeddings covering token position within the sequences. Since BERT only covers two segments in its input, we treat the odd segment as segment A and the even one as segment B. The sum of these three embeddings is fed to a bidirectional Transformer layer of BERT.

\begin{figure*}[!htbp]
\includegraphics[page=1,width=1\textwidth]{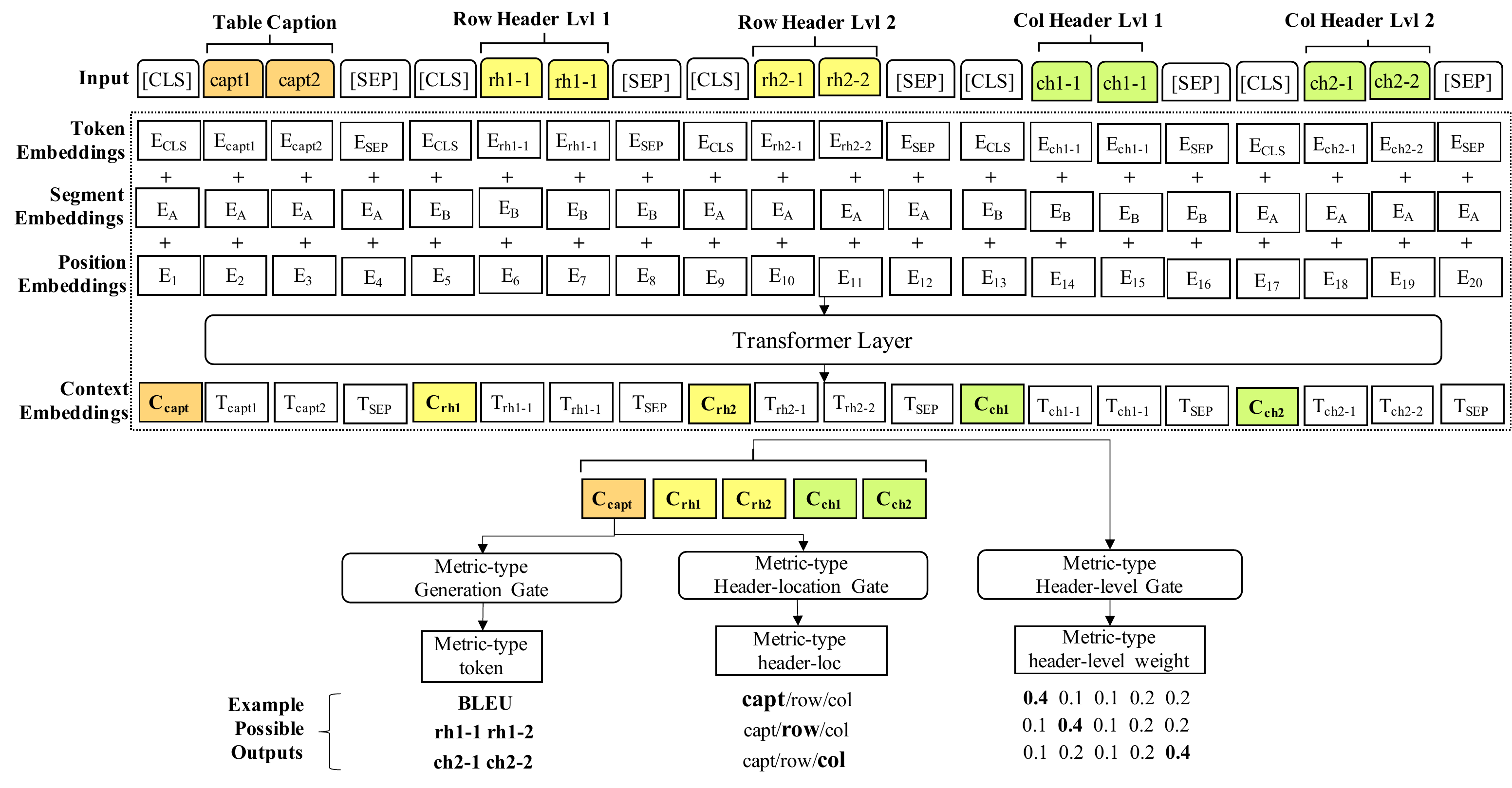}
    \caption{Architecture of proposed BERT-based model to identify metric-types in tables.}
    \label{arch-bert}
\end{figure*}

We use the token representations from the top hidden layers of the pre-trained Transformer as context embeddings. 
We assume the context vectors of each [CLS] token can represent the segment sequences better. As shown in Figure \ref{arch-bert}, we denote the input embedding as E, the final hidden vector of the [CLS] token for the $i^{th}$ input segment as $C_{i} \in \R_{H}$, and the final hidden vector for the $j^{th}$ input token as $T_{j} \in \R_{H}$.

We use a metric-type header-location gate and a metric-type header-level gate for metric-type location classification, and a metric-type generation gate to generate metric-type tokens from vocabulary covering out-of-header metric-types. Our BERT-based model architecture is shown in Figure \ref{arch-bert}. 

\paragraph{Metric-type header-location gates} We feed the first segment context $C_{1}$ to the softmax layer to obtain the metric-type header-location probability:
\begin{equation}
p_{hloc} = \text{softmax}(C_{1}).
\end{equation}

\paragraph{Metric-type header-level gates}
In our task, segments are used to represent the table section that is most related to metric-type. We incorporate the segment context $C_{i}$ to the sigmoid layer to obtain the probability of the metric-type being located in a specific header-level:
\begin{equation}
p_{hlevel_{i}} = \text{sigmoid}(C_{i}).
\end{equation}
The probabilities are then normalized to all segments as a weight score of the header-level:
\begin{equation}
\text{w}_{hlevel_{i}} = \frac{p_{hlevel_{i}}}{\sum^{n}_{i=1} p_{hlevel_{i}}}.
\end{equation}

\paragraph{Metric-type generation gates} 
We use a softmax function based on the first segment context $C_{1}$ to compute a probability distribution over the metric-type vocabulary:
\begin{equation}
P_{vocab}(w_{m}) = \text{softmax}(C_{1}).
\end{equation}

\paragraph{Learning objective} We combine all loss functions in the metric-type header-location, metric-type header-level, and metric-type generation gates:
\begin{multline}
\mathcal{L} = -((1-\alpha)(\sum _{c} z_{hloc} \log{p}_{hloc_{c}} +\\ \sum_{i=1}^{n}\log{\text{w}}_{{hlvl}_{i}}) + \alpha\log P_{vocab}(w_{m})),
\end{multline}
where $\alpha$ is the weight of the metric-type generation functions.

\section{Experimental Settings}
\subsection{Baseline Model}
We use two SVM classification models as baselines: a metric-type location prediction model and a metric-type token prediction model from the vocabulary of metric-types. We use tf.idf of the concatenation header name tokens for all levels as input representations in the first model and tf.idf of the caption tokens in the second one. 
We tuned hyperparameters of the SVM model and reported the best results.

\begin{table*}[!h]
\centering
\begin{tabular}{llllll}
\hline \textbf{Model} & $acc_{hloc}$ & $acc_{hlevel}$ & $acc_{m}^{sm}$ & $acc_{m\_token}^{sm}$ & $acc_{m\_token}^{ocm}$ \\ \hline
SVM & 81.48 & 82.96 & 67.41 & 69.83 & 69.83 \\
Pointer-Generator Supervised-Att (Glove) & 84.44 & 84.44 & 68.89 & 69.57 & 72.37 \\
Pointer-Generator Supervised-Att (BERT) & 67.41 & 51.11 & 45.93 & 33.66 & 36.61 \\
Pointer-Generator Supervised-Att (SciBERT) & 71.11 & 57.78 & 44.44 & 32.51 & 35.47 \\
Fine-tuned BERT & 91.11 & 90.37 & 74.81 & 77.50 & 80.46 \\
Fine-tuned SciBERT &  \textbf{93.33} & \textbf{93.33} & \textbf{79.26} & \textbf{81.61} & \textbf{85.06} \\
\hline
\end{tabular}
\caption{\label{exp-results} Test accuracies (\%) of different metric-type identification models. }
\end{table*}

\begin{table*}[h!]
\centering
\begin{tabular}{llllll}
\hline \textbf{Model} & $acc_{hloc}$ & $acc_{hlevel}$ & $acc_{m}^{sm}$ & $acc_{m\_token}^{sm}$ & $acc_{m\_token}^{ocm}$ \\ \hline
Pointer-Generator Supervised-Att (Glove) & 84.44 & 84.44 & 68.89 & 69.57 & 72.37 \\
- copy & 85.19 & 84.44 & 62.22 & 62.89 & 65.52 \\
- copy and generation & 82.96 & 82.22 & 56.30 & 54.35 & 56.98 \\
\hline
\end{tabular}
\caption{\label{ablation} Accuracy scores (\%) of ablation test of our pointer-generator-based model.}
\end{table*}

\begin{table*}[h!]
\centering
\begin{tabular}{llllll}
\hline \textbf{Model} & $acc_{hloc}$ & $acc_{hlevel}$ & $acc_{m}^{sm}$ & $acc_{m\_token}^{sm}$ & $acc_{m\_token}^{ocm}$ \\ \hline
Fine-tuned BERT & 91.11 & 90.37 & 74.81 & 77.50 & 80.46 \\
- segment embeddings & 87.41 & 87.41 & 72.59 & 75.70 & 78.00 \\
Fine-tuned SciBERT &  93.33 & 93.33 & 79.26 & 81.61 & 85.06 \\
- segment embeddings & 91.85 & 91.85 & 76.30 & 79.31 & 81.28 \\
\hline
\end{tabular}
\caption{\label{ablation2} Test accuracies (\%) of ablated BERT-based model without segment embeddings.}
\end{table*}

\begin{table}[t]
\centering
\begin{tabular}{@{}cc|rrrr@{}}
\multicolumn{1}{c}{} &\multicolumn{1}{c}{} &\multicolumn{4}{c}{Predicted} \\ 
\multicolumn{1}{c}{} & 
\multicolumn{1}{c|}{} & 
\multicolumn{1}{c}{LRow} & 
\multicolumn{1}{c}{LCol} &
\multicolumn{1}{c}{CCapt} & 
\multicolumn{1}{c}{Gen} \\ 
\cline{2-6}
\multirow[c]{4}{*}{\rotatebox[origin=tr]{90}{Actual}}
& LRow  & 0 & 0  & 0 & 0 \\[1.5ex]
& LCol  & 2 & 77 & 1 & 6 \\ [1.5ex]
& CCapt  & 1 & 6  & 16 & 3 \\[1.5ex]
& Gen& 0 & 5 & 0 & 18 \\ 
\cline{2-6}
\end{tabular}
    \caption{Confusion matrix of Pointer-Generator Supervised-Att (Glove) prediction.}
    \label{confusion-matrix}
\end{table}

\begin{table}[t]
\centering
\begin{tabular}{@{}cc|rrr@{}}
\multicolumn{1}{c}{} &\multicolumn{1}{c}{} &\multicolumn{3}{c}{Predicted} \\ 
\multicolumn{1}{c}{} & 
\multicolumn{1}{c|}{} & 
\multicolumn{1}{c}{LRow} & 
\multicolumn{1}{c}{LCol} &
\multicolumn{1}{c}{Gen} \\ 
\cline{2-5}
\multirow[c]{4}{*}{\rotatebox[origin=tr]{90}{Actual}}
& LRow  & 0 & 0  & 0  \\[1.5ex]
& LCol  & 0 & 80 & 6  \\ [1.5ex]
& Gen& 0   & 3 & 46  \\ 
\cline{2-5}
\end{tabular}
    \caption{Confusion matrix of Fine-tuned SciBERT prediction.}
    \label{confusion-matrix2}
\end{table} 

\subsection{Metrics Evaluation}
We use accuracy metrics to evaluate the metric-type location and generated metric-type tokens.

\paragraph{Metric-type location accuracy}
The target of the metric-type location prediction model is the metric-type located in the row headers, in the column headers, or not found in the headers. The accuracy of header-location ($acc_{hloc}$) is the rate of correct header-location predictions.

Since details about the metric-type location in the header-level are needed to identify metric-type token lists, we also compute the accuracy of metric-type header-level ($acc_{hlevel}$) using the ratio of correct header-level predictions to the total number of predictions.

\paragraph{Metric-type token accuracy} 
Let $\hat{m} = (\hat{w}_{m_{1}}, ..., \hat{w}_{m_{n}})$ denote the sequence of predicted metric-type tokens for $n_r$ rows or $n_c$ columns (depending on the header-location prediction), and $m = (w_{m_{1}}, ..., w_{m_{n}})$ denote the target ones: for example, $\hat{m} = (\text{f1, f1, f1})$ and $m = (\text{f-1, f-1, f-1})$. We calculate the metric-type token accuracy using string matching of all token lists in $\hat{m}$ and $m$:
\begin{equation}
acc_{m}^{sm} = \frac{\text{\# correct }\hat{m}}{\text{\# }\hat{m}},
\end{equation}
and string matching of each token pair $\hat{w}_{m_{i}}$ and $w_{m_{i}}$ in the token lists:
\begin{equation}
acc_{m\_token}^{sm} = \frac{\text{\# correct }\hat{w}_{m}}{\text{\# }\hat{w}_{m}}.
\end{equation}

To cover token prediction with an abbreviation, we compute the metric-type token accuracy based on the ordered character matching as follows:
\begin{equation}
acc_{m\_token}^{ocm} = \frac{d}{\text{\# }\hat{w}_{m}},
\end{equation}
where $d$ is the number of $\hat{w}_{m}$ 
whose characters are all found in $w_{m}$ in the same order. For example, the predicted token {\it RG1} is regarded as correct when the reference token is {\it ROUGE-1}.

\begin{table*}[h!]
\centering
\begin{tabular}{ll}
\hline 
Caption &  experimental results in exploring the shared syntactic order event detector\\
Header & rh level 1: [model, model, model] \\
 & rh level 2: [cl trans mlp, cl trans cnn, cl trans hbrid] \\
  & ch level 1: [pre., rec., f1] \\
Gold metric-type & [pre., rec., f1]\\
Predicted metric-type & [pre., rec., f1]\\
\hline 
Caption & results on the image and video datasets of sts task (pearson’s r × 100)\\
Header & rh level 1: [method, method, method]\\
 & rh level 2: [sts baseline, pivot, ours] \\
  & ch level 1: [ms-vid (2012), pascal (2014), pascal (2015)] \\
Gold metric-type & [r, r, r]\\
Predicted metric-type & [pearson’s, pearson’s, pearson’s]\\
\hline
\end{tabular}
\caption{\label{example-outputs} Example of table caption, headers, and predicted metric-type.}
\end{table*}

\subsection{Implementation Details}
We implemented our models using the AllenNLP library \cite{gardner-etal-2018-allennlp}. In our pointer-generator-based model, we used pre-trained word embeddings for initialization and two-layer BiLSTMs with 256 hidden sizes in both the caption and header-level encoders. We used dropout \cite{srivastava-hovy-2014-vector} with the probability $p = 0.1$. For optimization in the training phase, we used Adam as the optimizer with a batch size of $10$ and a learning rate of $3 \times 10^{-3}$ and $3 \times 10^{-5}$ in pointer-generator-based and BERT-based, respectively, with a slanted triangular schedule \cite{howard-ruder-2018-universal}. 
We trained the model for a maximum of $20$ epochs with early stopping on the validation set (patience of $10$) and set $\alpha$ to $0.5$. We used the original BERT and the domain-specific SciBERT uncased model to fine-tune our BERT-based model.

\section{Results}

\subsection{Experimental Results}

\paragraph{Model comparison} 
The performances of the proposed and baseline models are shown in Table ~\ref{exp-results}. We can see that the Pointer-Generator Supervised-Attention model initialized by Glove embeddings outperformed the baseline in predicting metric-type location. The accuracy of this model in the metric-type generation part mostly scored better than the baseline. However, the performances dropped significantly when the input was initialized by BERT as well as by SciBERT. BERT and SciBERT embeddings failed to guide our pointer-generator-based model in the metric-type identification task, especially in generating metric-type tokens.

The accuracy of our BERT-based model was significantly better than the others, achieving header-location and header-level prediction accuracy of more than 90\% and generation accuracy improvement of more than 7 points (\%). The fine-tuned BERT-based model using a domain-specific SciBERT led to significant performance gains in all metrics.

\paragraph{The effect of copy mechanism} We evaluated our pointer-generator-based model using an ablation test, as shown in Table~\ref{ablation}. As we can see, the performances of our generation model without a copy mechanism decreased. This demonstrates that incorporating the copy mechanism is beneficial in a metric-type token generation. Our model had the worst accuracy when it ran without a pointer-generator network since the location prediction model alone failed to handle out-of-header metric-types.

\paragraph{The effect of segment embeddings} Table \ref{ablation2} shows the effect of segment embeddings in our BERT-based model. The accuracies of Fine-tuned BERT and the SciBERT model without segment embeddings both decreased. This means that segment embeddings successfully discriminate header-level boundaries in the input representation of BERT-based models.

\subsection{Qualitative Analysis}
We analyzed the errors of our pointer-generator-based and fine-tuned SciBERT models by means of the confusion matrices shown in Tables \ref{confusion-matrix} and \ref{confusion-matrix2}. 
For better understanding, we simply define our outputs in the matrices as \enquote{LRow} for metric-type located in row headers, \enquote{LCol} for metric-type located in column headers, \enquote{CCapt} for metric-type copied from the caption, and \enquote{Gen} for metric-type generated from the vocabulary.
The matrix for the fine-tuned SciBERT model does not include the CCapt class since this model does not contain a copy mechanism.

As shown in the Table \ref{confusion-matrix}, the most correct classifications were for copying from the header (row and column), while the highest confusions were for copying from the caption and generation from the vocabulary. The accuracy of generating correct metric-type tokens from the vocabulary was 27.78\%, and the accuracy of copying a metric-type from the caption was 75\%. The copying mechanism contributes to a better performance than generation one.  

From the confusion matrix of the SciBERT-based model in Table \ref{confusion-matrix2}, we can see that the highest confusion was for copying from the header. We also computed the accuracy of generated metric-type tokens and found that just 58.7\% of the generated tokens were correct. 

We also investigated the errors in the predicted metric-type tokens. We found that the models tended to generate more generic metric-types; for example, they extracted \textit{score} as a prediction for the target \textit{accuracy}. On the other hand, our models generated similar terms to the ground truth metric-type, such as generating the metric-type \textit{pearson's} for the target \textit{r}. The examples are shown in Table \ref{example-outputs}.

\section{Conclusion}
In this work, we provided multi-level header numerical table datasets extracted from scientific papers consisting of header tables, captions, and metric-types. We introduced a metric-type identification task for multi-level header numerical tables, and proposed joint location prediction and generation models to solve the task. We have shown that our proposed model can identify metric-types from the multi-level header tables, both when the metric-types are included in the headers and when they are not. 

Our datasets only cover scientific papers in the computational linguistic domain. The generalization of our results beyond domain still remains an open question due to the difficulties of collecting comparable datasets in other domains without additional annotation by human experts. 

\section*{Acknowledgements}
Lya Hulliyyatus Suadaa is supported by the Indonesian Endowment Fund for Education (LPDP) and Okumura-Takamura-Funakoshi Laboratory, Tokyo Institute of Technology. 
This work is partially supported by JST PRESTO (Grant Number JPMJPR1655).
We thank the anonymous reviewers for helpful discussion of this work and comments on previous drafts of the paper.

\bibliography{anthology,eacl2021}
\bibliographystyle{acl_natbib}
\end{document}